\def\year{2021}\relax
\documentclass[letterpaper]{article} 
\usepackage{aaai21bak}  
\usepackage{times}  
\usepackage{helvet} 
\usepackage{courier}  
\usepackage[hyphens]{url}  
\usepackage{graphicx} 
\usepackage{natbib}  
\usepackage{caption} 
\usepackage{tabularx}
\usepackage{tikz}
\usepackage{mathrsfs}
\usepackage{multirow}
\usepackage{amsmath}
\usepackage{amsfonts}
\usepackage{circledsteps}
\usepackage[ruled]{algorithm2e}

\usepackage{pgfplots}

\nocopyright

\pgfplotsset{compat=1.16}

\pgfplotsset{
   legend entry/.initial=,
   every axis plot post/.code={%
       \pgfkeysgetvalue{/pgfplots/legend entry}\tempValue
       \ifx\tempValue\empty
           \pgfkeysalso{/pgfplots/forget plot}%
       \else
           \expandafter\addlegendentry\expandafter{\tempValue}%
       \fi
   },
}
\usetikzlibrary{decorations.pathreplacing}
\usetikzlibrary{patterns}
\usetikzlibrary{shapes,backgrounds}

\newsavebox{\tempbox}

\newlength{\starsize}
\newlength{\starspread}
\newcommand{\starcolor}{}
\tikzset{starsize/.code={\setlength{\starsize}{#1}},
         starspread/.code={\setlength{\starspread}{#1}},
         starcolor/.code={\renewcommand{\starcolor}{#1}}}
\tikzset{starsize=1mm,
         starspread=3mm,
         starcolor=black}
\pgfdeclarepatternformonly[\starspread,\starsize,\starcolor]
  {custom fivepointed stars}
  {\pgfpointorigin}
  {\pgfqpoint{\starspread}{\starspread}}
  {\pgfqpoint{\starspread}{\starspread}}
  {
   \pgftransformshift{\pgfqpoint{\starsize}{\starsize}}
   \pgfpathmoveto{\pgfqpointpolar{18}{\starsize}}
   \pgfpathlineto{\pgfqpointpolar{162}{\starsize}}
   \pgfpathlineto{\pgfqpointpolar{306}{\starsize}}
   \pgfpathlineto{\pgfqpointpolar{90}{\starsize}}
   \pgfpathlineto{\pgfqpointpolar{234}{\starsize}}
   \pgfpathclose%
   \pgfsetfillcolor{\starcolor}
   \pgfusepath{fill}
  }
\def\hexagonsize{5pt}
\pgfdeclarepatternformonly
  {hexagons}
  {\pgfpointorigin}
  {\pgfpoint{3*\hexagonsize}{0.866025*2*\hexagonsize}}
  {\pgfpoint{3*\hexagonsize}{0.866025*2*\hexagonsize}}
  {
   \pgfsetlinewidth{0.4pt}
   \pgftransformshift{\pgfpoint{0mm}{0.866025*\hexagonsize}}
   \pgfpathmoveto{\pgfpoint{0mm}{0mm}}
   \pgfpathlineto{\pgfpoint{0.5*\hexagonsize}{0mm}}
   \pgfpathlineto{\pgfpoint{\hexagonsize}{-0.866025*\hexagonsize}}
   \pgfpathlineto{\pgfpoint{2*\hexagonsize}{-0.866025*\hexagonsize}}
   \pgfpathlineto{\pgfpoint{2.5*\hexagonsize}{0mm}}
   \pgfpathlineto{\pgfpoint{3*\hexagonsize+0.2mm}{0mm}}
   \pgfpathmoveto{\pgfpoint{0.5*\hexagonsize}{0mm}}
   \pgfpathlineto{\pgfpoint{\hexagonsize}{0.866025*\hexagonsize}}
   \pgfpathlineto{\pgfpoint{2*\hexagonsize}{0.866025*\hexagonsize}}
   \pgfpathlineto{\pgfpoint{2.5*\hexagonsize}{0mm}}
   \pgfusepath{stroke}
  }

\newcommand{\newcite}[1]{\citeauthor{#1}~\shortcite{#1}}
\newlength{\hatchspread}
\newlength{\hatchthickness}
\newlength{\hatchshift}
\newcommand{\hatchcolor}{}
\tikzset{hatchspread/.code={\setlength{\hatchspread}{#1}},
         hatchthickness/.code={\setlength{\hatchthickness}{#1}},
         hatchshift/.code={\setlength{\hatchshift}{#1}},
         hatchcolor/.code={\renewcommand{\hatchcolor}{#1}}}
\tikzset{hatchspread=3pt,
         hatchthickness=0.4pt,
         hatchshift=0pt,
         hatchcolor=black}
\pgfdeclarepatternformonly[\hatchspread,\hatchthickness,\hatchshift,\hatchcolor]
   {custom north west lines}
   {\pgfqpoint{\dimexpr-2\hatchthickness}{\dimexpr-2\hatchthickness}}
   {\pgfqpoint{\dimexpr\hatchspread+2\hatchthickness}{\dimexpr\hatchspread+2\hatchthickness}}
   {\pgfqpoint{\dimexpr\hatchspread}{\dimexpr\hatchspread}}
   {
    \pgfsetlinewidth{\hatchthickness}
    \pgfpathmoveto{\pgfqpoint{0pt}{\dimexpr\hatchspread+\hatchshift}}
    \pgfpathlineto{\pgfqpoint{\dimexpr\hatchspread+0.15pt+\hatchshift}{-0.15pt}}
    \ifdim \hatchshift > 0pt
      \pgfpathmoveto{\pgfqpoint{0pt}{\hatchshift}}
      \pgfpathlineto{\pgfqpoint{\dimexpr0.15pt+\hatchshift}{-0.15pt}}
    \fi
    \pgfsetstrokecolor{\hatchcolor}
    \pgfusepath{stroke}
   }

\newcommand{\textbox}[1]
{\savebox{\tempbox}{#1}
 \ifdim\wd\tempbox<\columnwidth\relax
   \makebox[\columnwidth]{\usebox{\tempbox}}%
 \else
   \parbox{\columnwidth}{\raggedright #1}%
 \fi}
\begin{document}
%

\title{Semantics Altering Modifications \\for Evaluating Comprehension in Machine Reading}
\author{
    Viktor Schlegel, Goran Nenadic and Riza Batista-Navarro\\
}
\affiliations{


    Department of Computer Science, University of Manchester\\
    Manchester, United Kingdom\\
    \{viktor.schlegel, gnenadic, riza.batista\}@manchester.ac.uk

}
\maketitle
\begin{abstract}
Advances in NLP have yielded impressive results for the task of machine reading comprehension (MRC), with approaches having been reported to achieve performance comparable to that of humans. In this paper, we investigate whether state-of-the-art MRC models are able to correctly process Semantics Altering Modifications (SAM): linguistically-motivated phenomena that alter the semantics of a sentence while preserving most of its lexical surface form. We present a method to automatically generate and align challenge sets featuring original and altered examples. We further propose a novel evaluation methodology to correctly assess the capability of MRC systems to process these examples independent of the data they were optimised on, by discounting for effects introduced by domain shift. In a large-scale empirical study, we apply the methodology in order to evaluate extractive MRC models with regard to their capability to correctly process SAM-enriched data. We comprehensively cover 12 different state-of-the-art neural architecture configurations and four training datasets and find that -- despite their well-known remarkable performance -- optimised models consistently struggle to correctly process semantically altered data.






\end{abstract}
\section{Introduction}
Machine Reading Comprehension (MRC), also commonly referred to as Question Answering, is defined as finding the answer to a natural language question given an accompanying textual context. State-of-the-art approaches build upon large transformer-based language models \cite{Vaswani2017} that are optimised on large corpora in an unsupervised manner \cite{Devlin2018} and further fine-tuned on large crowd-sourced task-specific MRC datasets \cite{rajpurkar2016squad,Yang2018,Trischler2017}. They achieve remarkable performance, consistently outperforming human baselines on multiple reading comprehension and language understanding benchmarks \cite{Lan2020,Raffel2019}.

More recently, however, research on ``data biases'' in NLP suggests that these task-specific datasets exhibit various cues and spurious correlations between input and expected output \cite{gururangan2018annotation,Poliak2018,Schlegel2020a}. Indeed, data-driven approaches such as the state-of-the-art models (described above) that are optimised on these datasets learn to exploit these \cite{Jia2017,mccoy2019right}, thus circumventing the actual requirement to perform comprehension and understanding.

\begin{figure}[t]
    \centering
    \tikzset{every picture/.style={line width=0.75pt}} 
\tikzset{
mybrace/.style={decorate,decoration={brace,aspect=#1,amplitude=5pt}}
}
\begin{tikzpicture}[x=0.75pt,y=0.75pt,yscale=-1,xscale=1]

\draw   (0,0) -- (\columnwidth,0) -- (\columnwidth,240) -- (0, 240) -- cycle ;
\draw   (0,206) -- (\columnwidth,206) ;
\draw  [dashed] (10,33) -- (305,33);
\draw [mybrace=0.5] (10,140) -- (10,22);
\draw (4,0) node [anchor=north west][inner sep=0.5pt]  [font=\normalsize] [align=left] {\textbf{P:} \emph{\Circled{1} After the kickoff \textcolor[rgb]{0.29,0.56,0.89}{\underline{Naomi Daniel}}... }};
\draw (10,19) node [anchor=north west][inner sep=0pt]  [font=\normalsize] [align=center] { \textbf{(B) Original}: \emph{curled in}};
\draw (10,35) node [anchor=north west][inner sep=0pt]  [font=\normalsize] [align=center] { \textbf{(I1) Modal negation}: \emph{\emph{\textcolor[rgb]{0.82,0.01,0.11}{couldn't curl}} in}};
\draw (10,51) node [anchor=north west][inner sep=0pt]  [font=\normalsize] [align=center] { \textbf{(I2) Adverbial Modification}: \emph{\textcolor[rgb]{0.82,0.01,0.11}{\emph{almost}} curled in}};
\draw (10,67) node [anchor=north west][inner sep=0pt]  [font=\normalsize] [align=center] { \textbf{(I3) Implicit Negation}: \emph{\textcolor[rgb]{0.82,0.01,0.11}{\emph{was prevented from}} curling in}};
\draw (10,83) node [anchor=north west][inner sep=0pt]  [font=\normalsize] [align=center] { \textbf{(I4) Explicit Negation}: \emph{\textcolor[rgb]{0.82,0.01,0.11}{\emph{didn't succeed in}} curling in}};
\draw (10,99) node [anchor=north west][inner sep=0pt]  [font=\normalsize] [align=center] { \textbf{(I5) Polarity Reversing}: \emph{\textcolor[rgb]{0.82,0.01,0.11}{\emph{lacked the nerve to}} curl in}};
\draw (10,115) node [anchor=north west][inner sep=0pt]  [font=\normalsize] [align=left] { \textbf{(I6) Negated Polarity Preserving}: \textcolor[rgb]{0.82,0.01,0.11}{wouldn't find the} \\ \textcolor[rgb]{0.82,0.01,0.11}{opportunity to} \emph{curl in}};
\draw (5,146) node [anchor=north west][inner sep=0pt]  [font=\normalsize\itshape] [align=left] {\textbox{ ...a goal from 26 metres away following a decisive  coun- ter-attack. \Circled{2} Then \textcolor[rgb]{0.25,0.46,0.02}{\underline{Amanda Collins}} added more insult to the injury when she slotted in from 23 metres after Linda Burger’s soft clearance. [...]}};
\draw (0,210) node [anchor=north west][inner sep=0pt]  [font=\normalsize] [align=left] {{ \textbf{Q:} \emph{Who scored the farthest goal?}}};
\draw (0,226) node [anchor=north west][inner sep=0pt]  [font=\normalsize] [align=left] {{ \textbf{A:} \textcolor[rgb]{0.29,0.56,0.89}{Naomi Daniel}}};
\draw (135,226) node [anchor=north west][inner sep=0pt]  [font=\normalsize] [align=left] {{ \textbf{A with \textcolor[rgb]{0.82,0.01,0.11}{SAM}:} \textcolor[rgb]{0.25,0.46,0.02}{Amanda Collins}}};

\end{tikzpicture}
    \caption{Categories of SAM used in this paper with their implications on answering the given question. Modifying  the original \emph{``Baseline''} passage (B) by selecting any \emph{``Intervention''} category (I1)$-$(I6), or removing the first sentence (\emph{``Control''}) changes the correct answer from ``Naomi~Daniel'' located in sentence \Circled{1} to ``Amanda~Collins'' located in sentence \Circled{2}.}
    \label{fig:SAM}
\end{figure}

For a (simplified) example, consider the question \emph{``Who scored the farthest goal?''} illustrated in Figure~\ref{fig:SAM}. 
If a model is only exposed to examples where the accompanying passage contains sentences similar to ``X scored a goal from Y metres'' during training, a valid approximating decision based on this information could be similar to \emph{``select the name next to the largest number and the word goal''} without actually fully reading the passage. 

Alarmingly, conventional evaluation methodology where the dataset is split randomly into training and test data would not solve this issue. As both splits still stem from the same generative process (typically crowd-sourcing), the same types of cues are likely to exist in evaluation data, and a model can achieve high performance by relying on exploiting them. These and other problems suggest that the actual reading \emph{comprehension} of state-of-the-art MRC models is potentially over-estimated. 

In an attempt to present a more realistic estimate, we focus on the capability to correctly process \emph{Semantic Altering Modifications} (SAM): minimal modifications to the passage that change its meaning and therefore the expected answer. On the one hand, it is important to know whether these modifications are processed correctly by MRC models, as they drastically change the meaning, for example if ``X \emph{almost} scored a goal from Y metres'' then the goal effectively did not happen. On the other hand, distinguishing between original and modified examples is hard by relying on lexical cues only, as the modifications keep a similar lexical form. As a consequence, the simplified decision rule hypothesised above would not apply anymore. 


Manually curating evaluation data to incorporate SAM is expensive and requires expert knowledge; also, the process must be repeated for each dataset resource \cite{Gardner2020}. Automatically changing existing MRC data is not a feasible strategy either, as the effects of a change on the meaning of the passage cannot be traced through the process and will still need to be verified manually. Instead, in this paper we propose a novel methodology to generate SAM MRC challenge sets. We employ template-based natural language generation to maintain control over the presence of SAM and their effect onto the expected answer to a given question.

A problem that arises when evaluating models on challenge sets that were optimised on different training data, as it is the case in this paper, is the domain shift between training and evaluation data. For example, a model trained to retrieve answers from Wikipedia paragraphs might have never encountered a question involving comparing distances. In this case, wrong predictions on SAM examples cannot be attributed to the presence of SAM alone. To disentangle the effects introduced by the domain shift from the actual capability of correctly processing examples featuring SAM, we introduce a novel evaluation methodology with a corresponding metric, which we refer to as Domain Independent Consistency Evaluation or $DICE$. 
This allows us to precisely measure the capability of MRC models to process SAM of interest, and therefore, evaluate comprehension and language understanding that cannot be easily circumvented by relying on superficial cues. In a large-scale empirical study\footnote{Code and Supplementary Materials SM1, SM2 and SM3 can be retrieved from \mbox{\url{https://github.com/schlevik/sam}}}, we evaluate the performance of state-of-the-art transformer-based architectures optimised on multiple extractive MRC datasets. We find that while approaches based on larger language models tend to perform better, all investigated models struggle on the proposed challenge set, even after discounting for domain shift effects. 

\section{Semantics Altering Modifications}
\label{sec:sam}
The task of (extractive) Machine Reading Comprehension is formalised as follows: given a question $Q$ and a context $P$ consisting of words $p_0 \ldots p_n$, predict the start and end indices $s, e$ (where $s < e$) that constitute the answer span $A = p_s \ldots p_e$ in $P$. 
A Semantics Altering Modification (SAM) refers to the process of changing answer $A$ to $A' \neq A$ by applying a modification to the accompanying context $P$. The rationale is to create a new \emph{intervention} instance $(Q, P', A')$ that is lexically similar to the original but has a different meaning and therefore a different expected answer for the same question. Predicting both $A$ and $A'$ given the question and the respective passages becomes a more challenging task than predicting $A$ alone, since it requires correctly processing and distinguishing both examples. Due to their similarity, any simplifying heuristics inferred from training data are more likely to fail. 




Furthermore, this intuitive description aligns with one of the prevalent linguistic definitions of modifiers as ``an expression that combines with an unsaturated expression to form another unsaturated expression of the same [semantic] type'' \cite{McNally2002}. Particularly applicable to our scenario is the pragmatic or discourse-related view, specifically the distinction between modifiers that contribute to the content of a sentence with regard to a specific issue, and those that do not. In the context of MRC, the issue is whether the modification is relevant to finding the answer $A$ to the question $Q$.

The linguistics literature is rich in reporting phenomena conforming with this definition. In this paper we explore negation \cite{Morante2012}, (adverbial) restrictivity modification \cite[Sec.~6]{Tenny2000}, polarity reversing verbs and expressions \cite{Karttunen1971,Karttunen2012} and expressions of implicit negation \cite{Iyeiri2010}. The categories with representative examples are shown in Figure~\ref{fig:SAM} and labelled \emph{I1-I6}. They reflect our intuitive definition as they involve relatively small edits to the original context, by inserting between one and four words that belong to the most frequent parts of speech classes of the English language, i.e. adverbials, modals, verbs and nouns. Note, however, that this selection is non-exhaustive. Other linguistic phenomena such as privative adjectives \cite{Pavlick2016a}, noun phrase modification \cite{Stanovsky2016} or---if one were to expand the semantic types-based definition introduced above---corresponding discourse relations, such as Contrast or Negative Condition \cite{Prasad2008}, or morphological negation constitute further conceivable candidates. We leave it for future work to evaluate MRC on other types of SAM.





\section{Domain Independent Consistency Evaluation}
Consistency on ``contrastive sets'' \cite{Gardner2020} was recently proposed as a metric to evaluate the comprehension of NLP models beyond simplifying decision rules. A contrastive set is---similar to SAM---a collection of similar data points that exhibit minimal differences such that the expected prediction (e.g. answer for MRC) differs for each member. Consistency is then defined as the ratio of contrastive sets where the model yielded a correct prediction for all its members to the total number of contrastive sets. 

This notion requires that evaluation examples stem from the same generative process as the training data, making the process of finding contrastive sets dataset-dependent. If the processes are different however, as it is the case with training set-independent challenge sets, this difference can be a confounding factor for wrong predictions, i.e. a model might produce a wrong prediction because the input differs too much from its training data and not solely because it was not capable of solving the investigated phenomenon.
As we aim to establish an evaluation methodology independent of training data, 
we propose the following approach in order to rightfully attribute the capability to correctly process SAM even under domain shift.

We align each \emph{baseline} MRC instance consisting of question, expected answer and context triple $B_i = (Q_i, A_i, P_i)$ with an \emph{intervention} instance $I_i = (Q_i, A'_i, P'_i)$ s.t. $A'_i \neq A_i$. In practice, we achieve this by inserting a SAM in the sentence of $P_i$ that contains $A_i$ in order to obtain $P'_i$. We further align a \emph{control} instance where we completely remove the sentence that was modified in $P'_i$, i.e. $C_i = (Q_i, A'_i, P''_i)$. Thus, an \emph{aligned} instance consists of the triple $(B_i,I_i,C_i)$ sharing the question $Q$. The answer $A'$ is equivalent for both $I_i$ and $C_i$. $P, P'$ and $P''$ are shown in Figure~\ref{fig:SAM} by selecting original (B) for $P$, any of the alternatives (I1) through (I6) for $P'$ and completely removing the first sentence for $P''$. Formulating the problem in this way is similar to the Winograd Schema Challenge \cite{Levesque2012TheChallenge}.

The goal is to establish first, whether the model under evaluation ``understood'' the question and the accompanying context. Namely, if the model predicted $A_i$ and $A'_i$ correctly given $Q_i, P_i$ and $Q_i, P''_i$,  respectively, we conclude that the domain shift is not pivotal for the prediction performance of this particular instance, thus predicting the correct answer $A'_i$ for $I_i$ can be attributed to the model's capability to correctly process the SAM in $P'_i$. Conversely, if the model fails to predict $A'$ we assume that the reason for this is is its incapability to process SAM (for this instance), regardless of the domain shift. 

Initial experiments showed that models sometimes struggle to predict the exact span boundaries of the expected answer while retrieving the correct information in principle (e.g. predicting ``from 26 metres'' vs. the expected answer ``26 metres''). Therefore we relax the usual \emph{Exact Match} measure  $EM$ to establish the correctness of a prediction in the following way:  $rEM_{k}(\hat A,A) = 1$ if a $\hat A$ has at most $k$ words and $A$ is a substring of $\hat A$ and $0$ otherwise, where $\hat A = f_\theta(Q,P)$ is the answer prediction of an optimised MRC model $f_\theta$ given question $Q$ and context $P$.

The metric $DICE$ is the number of examples the model predicted correctly in their baseline, intervention and control version divided by the number of those the model predicted correctly for the baseline and control version. This notion reflects the ratio of those modified instances that the model processed correctly regardless of the domain shift thus further evaluating the model's reading comprehension. 
Formally, for a challenge set $\mathcal N = \mathcal{\{B, I, C\}}$ consisting of $N$ baseline, intervention and control examples, let 
\begin{align}
\label{eq:bic}
\begin{aligned}
\mathcal{B^+} &= \{i\ |\ rEM_k(f_\theta(Q_i, P_i), A_i) = 1\}_{i\in \{1 \ldots N\}} \\
\mathcal{I^+} &= \{i\ |\ rEM_k(f_\theta(Q_i, P'_i), A'_i) = 1\}_{i\in \{1 \ldots N\}} \\
\mathcal{C^+} &= \{i\ |\ rEM_k(f_\theta(Q_i, P''_i), A'_i) = 1\}_{i\in \{1 \ldots N\}}
\end{aligned}
\end{align}
denote the set of indices where an optimised model $f_\theta$ predicted a correct answer for baseline, intervention and control instances, respectively. Then

\begin{equation}
\label{eq:DICE}
    DICE(f_\theta) = \frac{|\mathcal{B^+} \cap \mathcal{I^+} \cap \mathcal{C^+}|}{|\mathcal{B^+} \cap \mathcal{C^+}|} \in [0,1] .
\end{equation}





An inherent limitation of challenge sets is that they bear negative predictive power only \cite{Feng2019}. Translated to our methodology, this means that while low $DICE$ scores hint at the fact that models circumvent comprehension, high scores do not warrant the opposite, as a model still might learn to exploit some simple decision rules in cases not covered by the challenge set. In other words, while necessary, the capability of distinguishing and correctly processing SAM examples is not sufficient to evaluate reading comprehension.

A limitation specific to our approach is that it depends on a model's capability to perform under domain shift, at least to some extent. If a model performs poorly because of insufficient generalisation beyond training data or if the training data are too different from that of the challenge set, the sizes of $\mathcal{B^+}, \mathcal{I^+}$ and $\mathcal{C^+}$ decrease and therefore variations due to chance have a larger contribution to the final result. Concretely, we found that if the question is not formulated in natural language, as is the case for \textsc{WikiHop} \citep{Welbl2018}, or the context does not consist of coherent sentences (with \textsc{SearchQA} \cite{Dunn2017} as an example) optimised models transfer poorly. Having a formalised notion of dataset similarity with respect to domain transfer for the task of MRC would help articulate the limitations and application scenarios of the proposed approach beyond pure empirical evidence.


\section{SAM Challenge Set Generation}
We now present the methodology for generating and modifying passages at scale. We aim to generate examples that require ``reasoning skills'' typically found in state-of-the-art MRC benchmarks \cite{Sugawara2017a,Schlegel2020}. Specifically, we choose to generate football match reports as it intuitively allows us to formulate questions that involve simple (e.g. \emph{``Who scored the first/last goal?''}) and more complex (e.g. \emph{``When was the second/second to last goal scored?''}) linear retrieval capabilities, bridging and understanding the temporality of events (e.g. \emph{``Who scored  before/after X was fouled?''}) as well as ordering (e.g. \emph{``What was the farthest/closest goal?''}) and comparing numbers and common properties (e.g. \emph{``Who assisted the earlier goal, X or Y?''}). Answer types for these questions are named entities (e.g. players) or numeric event attributes (e.g. time or distance).
\begin{figure}[tb]
    \centering
    \begin{tabularx}{1\columnwidth}{X}
        \hline
         \textbf{Selected Content Plan} \\
         \texttt{\footnotesize 1 (Order (Distance (Modified Goal) 0)} \\
         \texttt{\footnotesize 2 (Order (Distance (Just Goal) 1)} \\
         \texttt{\footnotesize Q (Argselect Max Goal Distance Actor)} \\
         \hline
         \textbf{Generated Events} \\
         \texttt{\footnotesize 1 \{actor: p4, distance: 26, {\color{red}mod: I2} \ldots\}} \\
         \texttt{\footnotesize 2 \{actor: p2, distance: 23 \ldots\}} \\
         \texttt{\footnotesize A: p4 A': p2} \\ 
         \hline
         \textbf{Chosen Templates (Simplified)} \\
         \texttt{\footnotesize 1 \%Con \#Actor {\color{red}@SAM} \$V.Goal \$PP.Distance\ldots} \\
         \texttt{\footnotesize 2 \#Actor \%Con she \$V.Score \$PP.Distance\ldots} \\
         \texttt{\footnotesize Q Who scored the farthest goal ?} \\
         \hline
         \textbf{Generated Text} \\
         $P$: After the kickoff Naomi Daniel curled in a goal \ldots \\
         $P'$: After the kickoff Naomi Daniel {\color{red} almost} curled in \ldots \\
         $P''$: Then Amanda Collins added more insult to the \ldots \\
         \hline
    \end{tabularx}
    \caption{Stages of the generative process that lead to the question answer and context in Figure~\ref{fig:SAM}. The \emph{Content Plan} describes the general constraints that the question type imposes on the \emph{Events} (both sentences must describe goal events, first sentence must contain SAM, distance attribute must be larger in the modified sentence). Appropriate \emph{Templates} are chosen randomly to realise the final Baseline $P$, Intervention $P'$ and Control $P''$ version of the passage.}
    \label{fig:generation}

\end{figure}

To generate passages and questions, we pursue a staged approach, common in Natural Language Generation \cite{Gatt2018}. 
Note that we choose a purely symbolic approach over statistical approaches in order to maintain full control over the resulting questions and passages as well as the implications of their modification for the task of retrieving the expected answer.
Our pipeline is exemplified in Figure~\ref{fig:generation} and consists of \emph{(1)} content determination and structuring, followed by \emph{(2)} content generation (as we generate the content from scratch) and finally \emph{(3)} lexicalisation and linguistic realisation combining templates and a generative grammar. 

\textbf{Content planning and generation:} The output of this stage is a structured report of events that occurred during a fictitious match, describing event properties such as actions, actors and time stamps. Furthermore each report is paired with a corresponding question, an indication of which event is to be modified, and the corresponding answers. The report is generated semi-randomly, as the requirement to generate instances with a \emph{meaningful} modification--i.e. actually changing the valid answer to the question--imposes constraints that depend on the type of the question. For example, for the retrieval type question \emph{"Who scored the farthest goal?"} the report must contain at least two events of the type ``goal'' and the distance attribute associated with the event to be modified must be larger. We generate events of the type ``goal'', which are the target of the generated questions and modifications,  and ``other'' that diverify the passages.  Furthermore, to prevent repetition, we ensure that the order of the types of events is unique for each report-question combination in the final set of generated reports. 

\textbf{Realisation:} For the sake of simplicity, we choose to represent each event with a single sentence, although it is possible to omit this constraint by using sentence aggregation techniques and multi-sentence templates. Given a structured event description, we randomly select a ``seed'' template suitable for the event type. Seed templates consist of variables that are further substituted by event properties and expressions generated by the grammar. Thereby, we distinguish between context-free and context-sensitive substitutions. For example \verb|$PP.Distance| in Figure~\ref{fig:generation} is substituted by a randomly generated prepositional phrase describing the distance (e.g. \emph{``from 26 metes away''}) regardless of its position in the final passage. \verb|%Con| in the same figure is substituted by an expression that connects to the previous sentence and depends on its content (e.g. \emph{``After the kick-off''} can only appear in the first sentence of the paragraph). We collect the templates and construct the grammar by combining various manual and automated measures described in SM: A in more detail. Similarly to the content generation, we ensure that the same template is not used more than once per report and the permutation of templates used to realise a report is unique in the final set of realised reports.








\textbf{Data description:}
The challenge set used in the experiments to evaluate MRC models trained on existing datasets consists of $4200$ aligned baseline, intervention and control examples generated using the above process. The modified intervention examples contain between one and three SAM from the six categories described earlier. Using $25$ ``seed'' templates and a generative grammar with $230$ production rules, we can realise an arbitrary event in $4.8\times 10^6$ lexically different ways; for a specific event the number is approx. $7.8\times 10^5$ on average (the difference is due to context-sensitive parts of the grammar). When fine-tuning MRC models on our generated data, we separate the seed templates in two distinct sets, in order to ensure that the models do not perform well by just memorising the templates. These template sets are used to generate a training (12000 instances) and an evaluation (2400 instances) set with aligned baseline, intervention and control instances.
All reports consist of six events and sentences, the average length of a realised passage is $174$ words, averaging $10.8$ distinct named entities and $6.9$ numbers as answer candidates.  

To estimate how realistic the generated MRC data is, we compare the paragraphs to the topically most similar MRC data: the NFL subset of the DROP dataset \cite{Dua2019}. We measure the following two metrics. \emph{Lexical Similarity} is the estimated Jaccard similarity between two paragraphs, i.e. the ratio of overlapping words, with lower scores indicating higher (lexical) diversity. As a rough estimate of \emph{Naturality}, we calculate the average of those sentence-level cohesion indices that are reported to correlate with human judgements of writing quality \cite{Crossley2016,Crossley2019}. For exact definitions and further results please consult SM1. The results are shown below:
\begin{center}
    \begin{tabularx}{0.92\columnwidth}{X c c}

\textbf{Data/Metric} & Lex. Similarity $\downarrow$ & Naturality $\uparrow$ \\
\hline 
SAM ($n=200$)  & $0.22$ &  $0.65$ \\
NFL ($n=188$) & $0.16$  &  $0.68$ \\
\hline
\end{tabularx}
\end{center}
While not quite reaching the reference data due to its template-based nature we conclude that the generated data is of sufficient quality for our purposes. 





\section{Experiments Setup}
Broadly, we address the following question: 
    \emph{How well does MRC perform on Semantic Altering Modifications?}

In this study we focus our investigations on extractive MRC where the question is in natural language, the context is one or more coherent paragraphs and the answer is a single continuous span to be found within the context. To that end we sample state-of-the-art (neural) MRC architectures and datasets and perform a comparative evaluation. Scores of models with the same architecture optimised on different data allow to compare how much these data enable models to learn to process SAM, while comparing models with different architectures optimised on the same data hints to which extent these architectures are able to obtain this capability from data.
Below we outline and motivate the choices of datasets and models used in the study. For further details on data preparation, model training and how we obtain predictions for evaluation, please consult SM2.







\textbf{Datasets:} 
We select the following datasets in an attempt to comprehensively cover various flavours of state-of-the-art MRC consistent with our definition above.
\begin{itemize}
    \item \textsc{SQuAD} \cite{rajpurkar2016squad} is a widely studied dataset where the human baseline is surpassed by the state of the art.
    \item \textsc{HotpotQA} \cite{Yang2018} in the ``distractor'' setting requires information synthesis from multiple passages in the context connected by a common entity or its property.
    \item \textsc{DROP} \cite{Dua2019} requires performing simple arithmetical tasks in order to predict the correct answer.
    \item \textsc{NewsQA} \cite{Trischler2017} contains questions that were created without having access to the provided context. The context is a news article, different from the other datasets where contexts are Wikipedia excerpts.
\end{itemize}

Similar to \newcite{Talmor2019}, we convert the datasets into the same format for comparability and to suit the task definition of extractive MRC. For \textsc{HotpotQA} we concatenate multiple passages into a single context and for \textsc{DROP} and \textsc{NewsQA} we only include examples where the question is answerable and the answer is a continuous span in the paragraph and refer to them as \textsc{DROP'} and \textsc{NewsQA'}, respectively. 

\textbf{Models:} The respective best-performing models on these datasets are employing a large transformer-based language model with a task-specific network on top. Note that we do not use architectures that make dataset-specific assumptions (e.g. ``Multi-hop'' for \textsc{HotpotQA}) in order to maintain comparability of the architectures across datasets. Instead, we employ a linear layer as the most generic form of the task-specific network \cite{Devlin2018}. Following common practice, we concatenate the question and context, and optimise the parameters of the linear layer together with those of the language model to minimise the cross-entropy loss between the predicted and expected start and end indices of the answer span (and the answer sequence for the generative model). 

We are interested in the effects of various improvements that were proposed for the original BERT transformer-based language model \cite{Devlin2018}. Concretely, we compare the effects of more training data and longer training for the language model (e.g. XLNet \cite{Yang2019a}, RoBERTa \cite{Liu2019c}), parameter sharing between layers of the transformer (e.g. ALBERT \cite{Lan2020}) and utilising a unifying sequence-to-sequence interface (e.g. BART \cite{Lewis2020}, T5 \cite{Raffel2019}) and reformulating extractive MRC as text generation conditioned on the question and passage. We evaluate models of different sizes, ranging from \texttt{base} (\texttt{small} for T5) to \texttt{large} (and \texttt{xl} and \texttt{xxl} for ALBERT). They describe specific configurations of the transformer architecture, such as the number of the self-attention layers and attention heads and the dimensionality of hidden vectors. For an in-depth discussion please refer to \newcite{Devlin2018} and the corresponding papers introducing the architectures. For comparison, we also include the non-transformer based BiDAF model \cite{Seo2017}. Finally, we train a model of the best performing architecture on a combination of all four datasets (\texttt{*-comb}) to investigate the effects of increasing training data diversity. For this, we sample and combine 22500 instances from all four datasets to obtain a training set that is similar in size to the others. The final selection consists of the models reported in Table~\ref{tab:results}.

\begin{table*}[!t]
    \centering
    \begin{tabularx}{1\textwidth}{X r | c c c c c c c c}
        & \textbf{Average} & \multicolumn{2}{c}{\textbf{\textsc{SQuAD}}} &  \multicolumn{2}{c}{\textbf{\textsc{HotpotQA}}} &  \multicolumn{2}{c}{\textbf{\textsc{NewsQA'}}} &  \multicolumn{2}{c}{\textbf{\textsc{DROP'}}} \\
        \textbf{Architecture} & \textbf{$DICE$} & \textbf{EM/F1} & \textbf{$DICE$} & \textbf{EM/F1} & \textbf{$DICE$} & \textbf{EM/F1} & \textbf{$DICE$} & \textbf{EM/F1} & \textbf{$DICE$} \\
        \hline
        \hline
 \texttt{bidaf}          & $11 \pm 3$ & $67.2/76.9$ & $12 \pm 4$ & $44.6/57.9$ & $4 \pm 3$  & $40.0/54.3$ & $13 \pm 5$ & $50.8/56.8$ & $18 \pm 12$ \\
\hline
 \texttt{bert-base}      & $13 \pm 2$ & $76.3/84.9$ & $13 \pm 3$ & $50.7/64.9$ & $17 \pm 4$ & $46.6/62.5$ & $13 \pm 3$ & $50.5/58.2$ & $10 \pm 3$  \\
 \texttt{bert-large}     & $15 \pm 2$ & $81.9/89.4$ & $15 \pm 3$ & $54.4/68.7$ & $14 \pm 3$ & $49.1/65.7$ & $14 \pm 4$ & $62.2/68.7$ & $16 \pm 3$  \\
 \texttt{roberta-base}   & $15 \pm 2$ & $82.4/89.9$ & $8 \pm 3$  & $51.9/66.4$ & $17 \pm 4$ & $50.8/66.9$ & $14 \pm 3$ & $63.5/69.3$ & $20 \pm 3$  \\
 \texttt{roberta-large}  & $18 \pm 1$ & $86.4/93.3$ & $16 \pm 3$ & $58.6/72.9$ & $21 \pm 3$ & $54.4/71.1$ & $15 \pm 3$ & $77.3/82.8$ & $20 \pm 2$  \\
 \texttt{albert-base}    & $14 \pm 2$ & $82.8/90.3$ & $10 \pm 3$ & $55.4/69.7$ & $17 \pm 3$ & $49.7/65.7$ & $11 \pm 3$ & $60.7/67.0$ & $18 \pm 4$  \\
 \texttt{albert-large}   & $16 \pm 1$ & $85.4/92.1$ & $18 \pm 3$ & $59.4/73.7$ & $12 \pm 2$ & $52.5/68.9$ & $17 \pm 3$ & $69.3/75.1$ & $18 \pm 3$  \\
 \texttt{albert-xl}  & $27 \pm 2$ & $87.1/93.5$ & $19 \pm 2$ & $62.4/76.2$ & $21 \pm 3$ & $54.2/70.4$ & $29 \pm 3$ & $76.4/81.8$ & $40 \pm 3$  \\
 \texttt{albert-xxl} & $27 \pm 1$ & $88.2/94.4$ & $29 \pm 2$ & $65.9/79.5$ & $29 \pm 3$ & $54.3/71.0$ & $25 \pm 3$ & $78.4/84.5$ & $23 \pm 2$  \\
 \texttt{t5-small}       & $10 \pm 1$ & $76.8/85.8$ & $13 \pm 3$ & $51.8/65.6$ & $10 \pm 3$ & $47.3/63.3$ & $8 \pm 2$  & $60.4/66.1$ & $10 \pm 3$  \\
 \texttt{t5-base}        & $16 \pm 1$ & $82.4/90.6$ & $16 \pm 3$ & $61.0/74.4$ & $20 \pm 3$ & $52.4/68.8$ & $14 \pm 3$ & $69.0/74.9$ & $15 \pm 2$  \\
 \texttt{t5-large}       & $20 \pm 1$ & $86.3/93.1$ & $21 \pm 2$ & $65.0/78.5$ & $29 \pm 3$ & $53.4/70.0$ & $16 \pm 3$ & $70.1/75.3$ & $8 \pm 2$   \\
\hline
 \texttt{average}        & $19 \pm 0$ & $76.4/83.2$ & $18 \pm 1$ & $53.1/65.9$ & $20 \pm 1$ & $47.1/62.1$ & $17 \pm 1$ & $61.5/67.0$ & $20 \pm 1$  \\
 \hline
 \texttt{albert-xl-comb}  & $20 \pm 2$ & $85.3/92.2$ & & $60.6/74.3$ &  & $53.6/70.4$ &  & $76.9/82.4$ &   \\
        \hline\hline
        \texttt{random} & {$5\pm0$} &\multicolumn{8}{r}{}\\
        \texttt{learned} & {$98\pm0$} &\multicolumn{8}{r}{}\\
        \hline 
     \end{tabularx}
    \caption{$DICE$ and EM/F1 score on the corresponding development sets of the evaluated models. Average $DICE$ scores are micro-averaged as it better shows the average performance on processing SAM examples while EM/F1 are macro-averaged as it reflects the average performance on the datasets (although the difference between both averaging methods is small in practice).}
    \label{tab:results}
\end{table*}

\textbf{Baselines:} 
We implement a \texttt{random} baseline that chooses an answer candidate from the pool of all named entities and numbers and an \texttt{informed} baseline that chooses randomly between all entities matching the expected answer type (e.g. person for ``Who'' questions). Finally, in order to investigate whether the proposed challenge can be solved in general, 
we train a \texttt{bert-base} model on 12000 aligned baseline and intervention instances, each. We refer to this baseline as \texttt{learned}. We train two more \texttt{bert-base} \emph{partial baselines}, \texttt{masked-q} and \texttt{masked-p} on the same data where, respectively, the question and passage tokens (except for answer candidates) are replaced by out-of-vocabulary tokens. Our motivation for doing this is to estimate the proportion of the challenge set that can be solved due to regularities in the data generation method, regardless of the realised lexical form to provide more context to the performance of the learned baseline. 
Finally, we estimate the \texttt{human} SAM performance by crowd-sourcing the manual annotation of 100 intervention examples. 

\section{Results and Discussion}
We present the main findings of our study here. For the obtained $DICE$ scores we report the error margin as a confidence interval at $\alpha=0.05$ using asymptotic normal approximation. Any comparisons between two $DICE$ scores reported in this section are statistically significant ($p<0.05$) as determined by performing the Fisher's exact test.




\begin{table}[b]
\centering
    \begin{tabularx}{0.9\columnwidth}{X c c c}

\textbf{Baseline} & $\mathcal{B}$ & $\mathcal{I}$ & $\mathcal{C}$ \\
\hline 
\texttt{learned} & $81 \pm 2$&  $79 \pm 2$  & $76 \pm 2$\\
\texttt{masked-q} & $20 \pm 2$&  $28 \pm 2$  & $26 \pm 1$ \\
\texttt{masked-p} &$29 \pm 1$&  $5 \pm 1$  &  $1 \pm 1$ \\
\texttt{random} & $6 \pm 1$&  $5 \pm 1$  & $8 \pm 1$ \\
\texttt{informed} & $14 \pm 1$&  $14 \pm 1$  & $26 \pm 2$ \\
\texttt{human} & -- &  $87 \pm 7$  & -- \\
\hline
\end{tabularx}
\caption{Percentage of correct predictions of the introduced baselines under the $rEM_5$ metric on aligned baseline $\mathcal{B}$, intervention $\mathcal{I}$ and control $\mathcal{C}$ sets.}
\label{tab:res-baseline}
\end{table}

\textbf{\emph{SAM is learnable.}} As expected, the learned baseline achieves high accuracy on our challenge set, with 81\% and 79\% correctly predicted instances for baseline and intervention examples, respectively, as seen in Table~\ref{tab:res-baseline}. The results are in line with similar experiments on Recognising Textual Entailment (RTE) and sentiment analysis tasks which involved aligned counterfactual training examples \cite{Kaushik2019a}. They suggest that neural networks are in fact capable of learning to recognise and correctly process examples with minimal yet meaningful differences such as SAM when explicitly optimised to do so. Some part of this performance is to be attributed to exploiting the regularity of the generation method rather than processing the realised text only, however, as the partial baselines perform better than the random baselines. This is further indicated by the slightly lower performance on the \emph{control} set, where due to deletion the number of context sentences is different compared to the baseline and intervention sets.

We note that the \texttt{learned} model does not reach 100\% EM score on this comparatively simple task, possibly due to the limited data diversity imposed by the templates. Using more templates and production rules and a bigger vocabulary will further enhance the diversity of the data.

\textbf{\emph{Pre-trained models struggle.}} Table~\ref{tab:results} reports the results of evaluating state-of-the-art MRC. Trained models struggle to succeed on our challenge set, with the best $DICE$ score of $40$ achieved by \texttt{albert-xlarge} when trained on \textsc{DROP'}.  
There is a log-linear correlation between the effective size of the language model (established by counting the shared parameters separately for each update per optimisation step) and the SAM performance with Spearman's $r~=~0.93$. Besides the model size, we do not find any contribution that leads to a considerable improvement in performance of \emph{practical} significance. 
We note that simply increasing the data diversity appears not beneficial, as the score of \texttt{albert-xl-comb} that was optimised on the combination of all four datasets is lower than the average score of the corresponding \texttt{albert-xl} model. Humans appear to have little issues to find the correct answer, with $87\% \pm 7\%$ of the intervention examples solved correctly. This provides the worst-case estimate for the human $DICE$ score, assuming all corresponding baseline and control examples would have been solved correctly (more details on how we establish the human baseline in SM3).

The easiest SAM category to process is \emph{I6: Explicit negation} with all optimised models scoring $26 \pm 1.4$ on average. Models struggle most with \emph{I2: Adverbial Modification}, with an average $DICE$ score of $14 \pm 1$ (see SM3 for breakdown by SAM category). A possible reason is that this category contains \emph{degree modifiers}, such as \emph{``almost''}. While they alter the semantics in the same way as other categories for our purposes, generally they act as a more nuanced modification (compare e.g. \emph{``almost''} with \emph{``didn't''}).
Finally, we note that the performance scales negatively with the number of SAM present in an example. The average $DICE$ score on instances with a single SAM is $23 \pm 0.9$, while on instances with the maximum of three SAM it is $16 \pm 0.8$ (and $19 \pm 1.0$ for two SAM). This is reasonable, as more SAM requires to process (and discard) more sentences, giving more opportunities to err.

We highlight that models optimised on \textsc{HotpotQA} and \textsc{DROP'} perform slightly better than models optimised on \textsc{SQuAD} and \textsc{NewsQA'} (on average $20\%$ vs $18\%$ and $17\%$, respectively). This suggests that exposing models to training data that require more complex (e.g. ``multihop'' and arithmetic) reasoning to deduce the answer, as opposed to simple  answer retrieval based on predicate-argument structure \cite{Schlegel2020}, has a positive effect on distinguishing and correctly processing lexically similar yet semantically different instances.







\textbf{\emph{Small improvements can be important.}} Our results indicate that small differences at the higher end of the performance spectrum can be of practical significance for the comprehension of challenging examples, such as SAM. Taking \texttt{albert} as an example, the relative performance improvements between the \texttt{base} and \texttt{xxlarge} model when (macro) averaged over the EM and F1 scores on the corresponding development sets are $15 \%$ and $13 \%$, respectively, while the relative difference in average $DICE$ score is $93 \%$. This is likely due to a share of ``easy'' examples in MRC evaluation data \cite{Sugawara2018} that artificially bloat the (lower-end) performance scores to an extent. 









\textbf{\emph{Meaningful training examples are missing.}} One possible explanation for low scores could be that the models simply never encountered the expressions we use to modify the passages and thus fail to correctly process them. To investigate this claim we count the occurrences of the expressions of the worst performing category overall, \emph{I2: Adverbial Modification}. The expressions appear in $5\%, 14\%, 5\%$ and $22\%$ of the training passages of \textsc{SQuAD}, \textsc{HotpotQA}, \textsc{DROP'} and \textsc{NewsQA'} respectively, showing that models do encounter them during task-specific fine-tuning (not to mention during the language-model pre-training). It is more likely that the datasets lack examples where these expressions affect the search for the expected answer in a meaningful way \cite{Schlegel2020}.
In fact, after sampling $400$ passages and $647$ corresponding questions (100 passages from each dataset) where the expression occurs within $100$ characters of the expected answer, and manually annotating whether the modification would yield a different answer, we find only one such case which we can thus consider as a naturally occurring SAM.
Worse yet, in $4\%$ of the cases the expected answer ignores the presence of the SAM.
This lends further credibility to the hypothesis that current MRC struggles at distinguishing examples with minimal yet meaningful changes such as SAM, if not explicitly incentivised during training. For more details on this annotation task, see SM3.

An analysis of models' errors suggests a similar conclusion: examining wrong intervention predictions for those cases where the answers for baseline and control were predicted correctly, we find that in $82\% \pm 1\%$ of those cases the models predict the baseline answer. Models thus tend to ignore SAM, rather than being ``confused'' by their presence (as if never encountered during training) and predicting a different incorrect answer. 

\section{Related Work} Systematically modified MRC data can be obtained by rewriting questions using rule-based approaches \cite{Ribeiro2018,Ribeiro2019} or appending distracting sentences, e.g. by paraphrasing the question \cite{Jia2017,Wang2018}, or whole documents \cite{Jiang2019} to the context. Adversarial approaches with the aim to ``fool'' the evaluated model, e.g. by applying context perturbations \cite{Si2020} fall into this category as well. These approaches differ from ours, however, in that they aim to preserve the semantics of the modified example, therefore the expected answer is unchanged. But the findings are similar: models struggle to capture the semantic equivalence of examples after modification, and rely on lexical overlap between question and passage \cite{Jia2017}. Our approach explores a complementary direction by generating semantically altered passages.

Using rule-based NLG techniques for controlled generation of MRC data was employed to obtain stories \cite{Weston2015} that aim to evaluate the learnability of specific reasoning types, such as inductive reasoning or entity tracking. Further examples are \texttt{TextWorld} \cite{Cote2018}, an environment for text-based role playing games with a dataset where the task is to answer a question by interactively exploring the world \cite{Yuan2019a} and extending datasets with unanswerable questions  \cite{Nakanishi2018}. Similar to our approach, these generation methods rely on symbolic approaches to maintain control over the semantics of the data.

Beyond MRC, artificially constructed challenge sets were established with the aim to evaluate specific phenomena of interest, particularly for the RTE task. Challenge sets were proposed to investigate neural RTE models for their capabilities to perform logic reasoning \cite{Richardson2019} and lexical inference \cite{Glockner2018}, or understanding language compositionality \cite{Nie2019a,Geiger2019}. 

\section{Conclusion}
We introduce a novel methodology for evaluating the reading comprehension of MRC models by observing their capability to distinguish and correctly process lexically similar yet semantically different input. We discuss lingustic phenomena that act as Semantic Altering Modifications and present a methodology to automatically generate and meaningfully modify MRC evaluation data. In an empirical study, we show that while the capability to process SAM correctly is learnable in principle, state-of-the-art MRC architectures optimised on various MRC training data struggle to do so. We conclude that one of the key reasons for this is the lack of challenging SAM examples in the corresponding datasets. 

Future work will include the search for and evaluation on further linguistic phenomena suitable for the purpose of SAM, expanding the study from strictly extractive MRC to other formulations such as generative or multiple-choice MRC, and collecting a large-scale natural language MRC dataset featuring aligned SAM examples (e.g. via crowd-sourcing) in order to investigate the impact on the robustness of neural models when exposed to those examples during training.
\section{Acknowledgements}
The authors would like to thank the anonymous reviewers for valuable suggestions as well as Alessio Sarullo for many fruitful discussions and help with the design of the experiments. The computation-heavy aspects of this paper were made possible due to access to the Computational Shared Facility at The University of Manchester, for which the authors are grateful.
\bibliography{short-refs}

\end{document}